\documentclass{nature}

\usepackage{etoolbox}
\usepackage{sansmath}
\usepackage{microtype}
\usepackage{algorithm}
\usepackage{algpseudocode}
\usepackage{graphicx}
\usepackage{verbatim}
\usepackage{booktabs}
\usepackage[flushleft]{threeparttable}
\usepackage{multirow}
\usepackage{color,soul}
\usepackage{caption}
\usepackage{subcaption}
\usepackage{todonotes}
\usepackage[hyphens]{url}  
\usepackage{selectp} 
\usepackage{fancyhdr}

\usepackage{booktabs}
\usepackage{caption}
\usepackage{subcaption}
\usepackage{amsmath}
\usepackage{mathtools}

\usepackage[breaklinks=true,colorlinks,allcolors=blue]{hyperref}
\usepackage[capitalise]{cleveref} 
\usepackage[modulo, switch]{lineno}
\usepackage{tikz}

\usepackage{authblk}
\usepackage{textcomp}
\usepackage{subcaption}
\usepackage{tabularx}
\usepackage{caption}
\newcommand*{\balancecolsandclearpage}{
  \close@column@grid
  \cleardoublepage
  \twocolumngrid
}

\makeatletter 
\g@addto@macro\TPT@defaults{\scriptsize}
\makeatother

\title{No Fair Lunch: A Causal Perspective on Dataset Bias in Machine Learning for Medical Imaging}

\author[1*]{Charles Jones}
\author[2]{Daniel C. Castro}
\author[1]{Fabio De Sousa Ribeiro}
\author[2]{Ozan Oktay}
\author[3]{Melissa McCradden}
\author[1*]{Ben Glocker}

\affil[1]{Department of Computing, Imperial College London, United Kingdom}
\affil[2]{Microsoft Health Futures, Cambridge, United Kingdom}
\affil[3]{The Hospital for Sick Children, Canada\vspace{2mm}
}

\affil[*]{\small Corresponding authors: \{charles.jones17,b.glocker\}@imperial.ac.uk \vspace{3mm}}

\begin{document}

\twocolumn[{
\begin{@twocolumnfalse}

\maketitle

\begin{abstract}
As machine learning methods gain prominence within clinical decision-making, addressing fairness concerns becomes increasingly urgent. Despite considerable work dedicated to detecting and ameliorating algorithmic bias, today's methods are deficient with potentially harmful consequences. Our causal perspective sheds new light on algorithmic bias, highlighting how different sources of dataset bias may appear indistinguishable yet require substantially different mitigation strategies. We introduce three families of causal bias mechanisms stemming from disparities in prevalence, presentation, and annotation. Our causal analysis underscores how current mitigation methods tackle only a narrow and often unrealistic subset of scenarios. We provide a practical three-step framework for reasoning about fairness in medical imaging, supporting the development of safe and equitable AI prediction models.
\end{abstract}

\end{@twocolumnfalse}}]

\section{Introduction}

Machine learning (ML) algorithms for medical image analysis are rapidly becoming powerful tools for supporting high-stakes clinical decision-making. However, these algorithms can reflect or amplify problematic biases present in their training data \cite{charImplementingMachineLearning2018,obermeyerDissectingRacialBias2019,wiensNoHarmRoadmap2019,buolamwiniGenderShadesIntersectional2018}. While ML methods have potential to improve outcomes by improving diagnostic accuracy and throughput, they often generalise poorly across clinical environments \cite{beedeHumanCenteredEvaluationDeep2020} and may exhibit algorithmic bias leading to worse performance in underrepresented populations \cite{seyyed-kalantariCheXclusionFairnessGaps2020,seyyed-kalantariUnderdiagnosisBiasArtificial2021,mamaryRaceGenderDisparities2018,oakden-raynerHiddenStratificationCauses2020,gianfrancescoPotentialBiasesMachine2018,larrazabalGenderImbalanceMedical2020}. In recent years, many methods for mitigating algorithmic bias in image analysis have been proposed \cite{wangFairnessVisualRecognition2020,zietlowLevelingComputerVision2022,alviTurningBlindEye2018,kimLearningNotLearn2019,madrasLearningAdversariallyFair2018,edwardsCensoringRepresentationsAdversary2016,ramaswamyFairAttributeClassification2021,wangRacialFacesWild2019,hendricksWomenAlsoSnowboard2018,liREPAIRRemovingRepresentation2019,quadriantoDiscoveringFairRepresentations2019,wangBalancedDatasetsAre2019}, however, it remains unclear when each method is most appropriate or even valid to employ \cite{corbett-daviesMeasureMismeasureFairness2018,friedlerComparativeStudyFairnessenhancing2019,zongMEDFAIRBenchmarkingFairness2023}. Bias mitigation methods may show convincing results on one benchmark yet appear useless or even harmful on others \cite{zietlowLevelingComputerVision2022}, raising the question: how should we effectively tackle bias in ML for medical imaging?

In this perspective, we discuss how the language of causality can illuminate the study of fairness. Inspired by causal work in dataset shift \cite{castroCausalityMattersMedical2020,subbaswamyDevelopmentDeploymentDataset2020,subbaswamyCounterfactualNormalizationProactively2018,subbaswamyPreventingFailuresDue2019,huangDistributionShiftMining2017} and domain adaptation \cite{yueTransportingCausalMechanisms2021,zhangMultiSourceDomainAdaptation2015,magliacaneDomainAdaptationUsing2018}, we argue that understanding the mechanisms of dataset bias which may \emph{cause} predictive models to become unfair is critical for reasoning about what steps to take in practice. We discuss how bias mitigation techniques may successfully combat one bias mechanism but be theoretically futile against others. We identify three key families of clinically relevant bias mechanisms: (i) \emph{prevalence disparities}, (ii) \emph{presentation disparities}, and (iii) \emph{annotation disparities}. We highlight how careful analysis of the underlying characteristics of these mechanisms is critical for informing the selection of appropriate methods and metrics. Our causal considerations are intended to help guide future research in the field.

\section{Dataset Bias in Imaging}

We consider image classification problems where we aim to learn a fair disease classifier, given a potentially biased training dataset of images and targets (e.g. class labels). Where much work focuses on algorithmic bias \cite{chenAlgorithmicFairnessArtificial2023}, concerning errors introduced by imperfect models, we take a step back and focus on dataset bias. From a machine learning perspective, dataset bias would be reflected even by a `perfect' model with no error on its biased training dataset. By focusing on what such a model would learn from a given dataset, we can use tools from causal reasoning to study how the underlying mapping between images and targets shifts across groups and settings. This is especially relevant in today's paradigm of high-capacity deep learning models, which often achieve near-zero training error when trained with the standard approach of empirical risk minimization (ERM)  \cite{vapnikOverviewStatisticalLearning1999}. Throughout this article, we take a graphical approach to causality, relying on the principles explained in our brief primer in Appendix \ref{sec:causalprimer}. We refer readers looking for further introductory material to Peters et al. \cite{petersElementsCausalInference2017} and to Pearl \cite{pearlCausalityModelsReasoning2011}.

\subsection*{Causality and sensitive information}

We begin with a causal formulation of the image classification problem, examining the data-generating process behind medical datasets. In medical image analysis, we deal with medical scans $X$ from patients with an underlying condition $Z$, causing pathological structures to be visible in their scans (${Z \rightarrow X}$, read as `$Z$ \emph{causes} $X$'). We wish to learn a model which may classify the condition of unlabelled images in deployment. However, since $Z$ is not directly observable, we must train our models to predict a proxy target $Y$, collected as part of the training dataset. For didactic clarity, our examples in this article primarily focus on tasks where $Y$ derives from $Z$, such as confirmed diagnoses or patient outcomes. We may view this setup as generalising the common assumption of anticausal classification \cite{scholkopfCausalAnticausalLearning2012,castroCausalityMattersMedical2020}. Where anticausal tasks traditionally assume that the target is a gold standard for the underlying condition (${Y \coloneqq Z}$), our formulation allows for the more general case where there may be label noise (${Y \leftarrow Z}$). Finally, we include a selection node $S$ in our diagrams to consider possible selection biases, where all observed data points are conditioned on $S$. When selection is random, we may omit $S$ for simplicity.

\begin{figure}[ht]
    \centering
    \includegraphics[width=\linewidth]{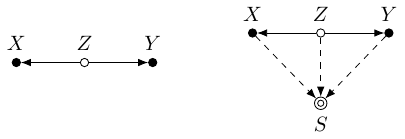}
    \caption{Basic causal structures of medical imaging tasks. An underlying condition $Z$ influences the image $X$ and label $Y$ for each individual. Selection $S$ may be random (left) or dependent on any combination of $\{X, Y, Z \}$ (right). In all cases, we wish to learn a model from observed data to predict $P(Y \mid X)$. Unfilled nodes represent unobserved variables, and concentric nodes represent selection variables, which are conditioned upon. We use this convention throughout this section.}
    \label{fig:causalmodel}
\end{figure}

\begin{figure}[ht]
    \centering
    \includegraphics[width=\linewidth]{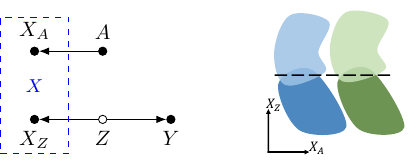}
    \caption{Basic causal structure of a no-bias dataset (left). We include a 2D illustration of the no-bias observational distribution (right). Green and blue clusters represent different subgroups, with light and dark clusters representing disease and no disease, respectively. Notice that the optimal decision boundary (dashed line) does not depend on $X_A$.}
    \label{fig:nobiasdag}
\end{figure}

The defining theme of fair image analysis is that images contain sensitive information about individuals, which models may learn to exploit inappropriately. In group fairness, this information encodes membership of a clinically or socially relevant population subgroup, denoted by a sensitive attribute $A$ (such as self-reported race, biological sex, age, socioeconomic status etc.). It is thus helpful for fairness analysis to construct our causal diagram with the image decomposed into two causal factors: $X_Z$, representing the pathological structures directly caused by the disease (${Z \rightarrow X_Z}$), and $X_A$, features that encode subgroup-related sensitive information (${A \rightarrow X_A}$)\footnote{We define sensitive information as any image features that are predictive of the sensitive attribute. For example, skin pigmentation may be sensitive information if self-reported race is the sensitive attribute. If geographical location is the attribute and patients in different hospitals are scanned by different machines, artefacts caused by equipment settings may be considered sensitive.}. When these causal factors are independent, $X_A$ provides no information about the disease, so it is not useful for the classification task. This constraint is expressed as $P(Y \mid X_Z, X_A, S) = P(Y \mid X_Z, S)$, equivalent to the conditional independence statement $Y \perp X_A \mid \{X_Z, S\}$. We call this the `no-bias' criterion. With a no-bias dataset, a poorly trained model may exhibit performance disparities across groups, but this would result from algorithmic bias, not dataset bias. In Fig. \ref{fig:nobiasdag}, we illustrate the simplest structure that satisfies the no-bias criterion, where the independence relationship may be derived through d-separation \cite{vermaCausalNetworksSemantics1990,pearlIdentifyingIndependenciesCausal1996} (explained in Appendix \ref{sec:causalprimer}).

\begin{figure}[ht]
    \centering
    \includegraphics[width=\linewidth]{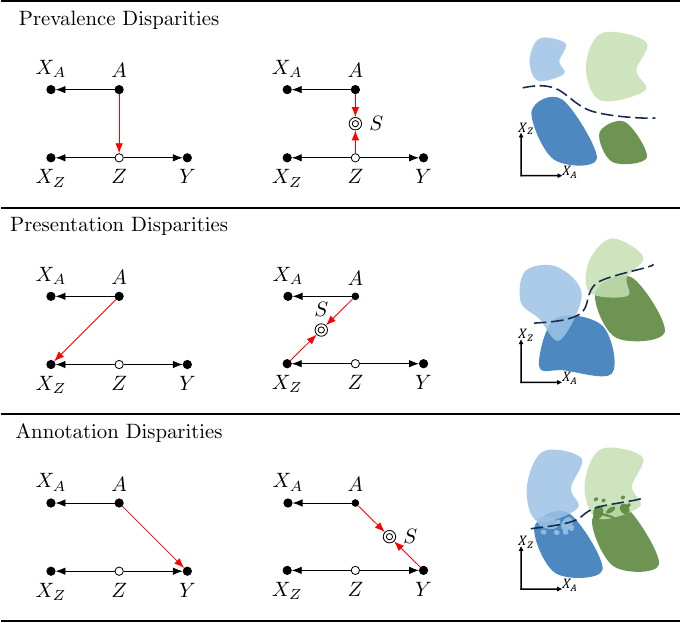}
    \caption{Causal structures of dataset bias in medical imaging. The relevant causal pathways for each are highlighted in red. Prevalence disparities involve a path between the sensitive attribute and the disease prevalence. Presentation disparities involve a path between the sensitive attribute and the disease presentation. Annotation disparities involve a path between the sensitive attribute and the annotation policy. These disparities may occur due to direct causal links (left diagrams) or collider biases from selection effects (middle diagrams). Each bias causes the optimal decision boundary to depend on sensitive information (right diagrams).}
    \label{fig:basicdags}
\end{figure}

In practice, we often see that models trained with ERM to predict $Y$ also acquire the ability to predict $A$ \cite{glockerAlgorithmicEncodingProtected2023,gichoyaAIRecognitionPatient2022,jonesRoleSubgroupSeparability2023}. This occurs when the no-bias criterion does not hold, and thus exploiting sensitive information improves task performance on the biased training dataset. We wish to represent such situations with our causal diagram. By applying d-separation to Fig. \ref{fig:nobiasdag}, we can see that any causal pathway linking $A$ to $Z$, $X_Z$, or $Y$ provides a connecting path between $X_A$ and $Y$ and hence violates the no-bias criterion. These are the core causal mechanisms behind dataset bias in medical imaging\footnote{While there may be additional settings where the no-bias criterion is violated by unobserved confounding $U$ between $X_A$ and $X_Z$, notice that this subgraph (${X_A \leftarrow U \rightarrow X_Z}$) is near-identical to the confounding in our formulation of presentation disparities. We thus expect insights on presentation disparities to be similarly applicable in settings where $X_A$ and $X_Z$ are intrinsically entangled due to causal inter-relationships. Detection and mitigation, however, may be more challenging when $U$ is unobserved, unlike $A$.} and are shown in Fig. \ref{fig:basicdags}. We refer to them as \emph{prevalence disparities}, \emph{presentation disparities}, and \emph{annotation disparities}, respectively.

\subsection*{Fair and unfair causal pathways}

Although dataset bias induces models to exploit sensitive information, this is not necessarily inappropriate in all cases. If a legitimate biological difference exists between groups, sensitive information may be relevant for disease prediction\cite{mccraddenWhatFairFair2023}. In this case, we would want a trained model to capture this information by learning group-specific disease mechanisms. In contrast, models may unfairly reflect spurious correlations caused by historical biases in healthcare provision or the diagnostic process. To account for both possibilities, we must distinguish between fair and unfair causal pathways \cite{chiappaPathSpecificCounterfactualFairness2019}. Fair causal pathways contain information our model should use to inform predictions, whereas unfair pathways should be ignored or mitigated by a learning algorithm. 

For real-world datasets, determining whether a causal pathway is fair or unfair is challenging and requires specific knowledge, including the ethical considerations of the particular application domain\cite{mccraddenWhatFairFair2023}. A simplistic guide is to imagine an unbiased deployment setting and consider whether we would expect a particular causal pathway to remain present. If not, the path is likely unfair. For example, a historic labelling policy may cause some groups to be underdiagnosed in training data -- if this effect is not expected to persist in deployment, we say that the policy is unfair. This heuristic allows us to convert any fairness problem into a dataset shift problem. A fair model in this context is one with perfect classification performance when deployed on a (potentially hypothetical) dataset with all unfair causal pathways removed. This problem framing helps to simplify and clarify the consideration of fairness methods and metrics. 

\subsection*{Measuring and mitigating bias}
 
Group fairness metrics aim to quantify algorithmic bias by measuring how classifier properties differ across subgroups. However, they are difficult to interpret and are often mutually incompatible \cite{friedlerImPossibilityFairness2016}. With our causal formulation, we may recontextualise group fairness metrics as (necessary but not sufficient) measures of properties we expect an unbiased prediction model to have. When fairness metrics are incompatible, we may often view each metric as assuming a different form of the hypothetical unbiased dataset. For example, when all causal pathways are considered fair, the unbiased deployment setting may be assumed to be independent and identically distributed (i.i.d.) to the training data. In this case, so-called `bias-preserving' metrics \cite{wachterBiasPreservationMachine2021}, such as equal opportunity \cite{hardtEqualityOpportunitySupervised2016}, should be favoured, and fairness is aligned with maximising performance at training time. In contrast, when our dataset contains unfair pathways, our hypothetical unbiased setting is not i.i.d. to the training dataset, so train-time performance may be misleading, and we should favour `bias-transforming' \cite{wachterBiasPreservationMachine2021} metrics such as demographic parity \cite{zemelLearningFairRepresentations2013}. Causal analysis such as this may help practitioners to understand better the assumptions involved in choosing fairness metrics, as well as the illusive concept of fairness--accuracy tradeoffs under different sources of dataset bias \cite{duttaThereTradeOffFairness2020,wickUnlockingFairnessTradeoff2019}. We refer readers to Ple\v{c}ko and Bareinboim \cite{plecko2022causal} for a more detailed causal examination of fairness metrics.

One advantage of framing the bias mitigation problem as one of generalisation to an unbiased dataset is that we can use causal transportability theory \cite{maoCausalTransportabilityVisual2022,pearlTransportabilityCausalStatistical2011,jiangInvariantTransportableRepresentations2022} to investigate the circumstances under which we may expect a model trained on the biased dataset to be appropriate for the unbiased deployment setting. Revisiting the setting where all causal pathways are considered fair, empirical risk minimisation (ERM) is an appropriate learning strategy, as we wish to maximise performance on an i.i.d. deployment setting \cite{vapnikOverviewStatisticalLearning1999}. Conversely, ERM is inappropriate in cases of unfair presentation disparities and unfair annotation disparities, as the underlying mapping between disease features and targets is expected to shift from training to deployment. The case of unfair prevalence disparities is particularly interesting, as the disparity may disappear in the limit of infinite data or with targeted data collection. However, in practice, the class imbalance is likely to lead to ERM models becoming miscalibrated, and there may be further issues if the observed data does not cover the support of the unbiased distribution \cite{castroCausalityMattersMedical2020}. Detailed causal transportability analysis of further bias mitigation methods, such as adversarial training \cite{alviTurningBlindEye2018,kimLearningNotLearn2019,madrasLearningAdversariallyFair2018,edwardsCensoringRepresentationsAdversary2016}, is beyond the scope of this article but is fertile ground for future work. We briefly discuss the applicability of prominent methods from the literature in Section \ref{sec:mechanisms}, emphasising that no method may effectively tackle all biases. 

\subsection*{No fair lunch}

When viewing a fair model as one which generalises from a biased training dataset to an unbiased deployment one, a simple problem reveals itself: no model may attain perfect performance over all possible deployment settings \cite{wolpertNoFreeLunch1997}. While we've demonstrated in anticausal settings how different biases may require different mitigation strategies, the general case is even more challenging. For any given training dataset, we could always construct opposing causal models which are equally compatible with the observed data \cite{hollandStatisticsCausalInference1986} and would render the same model or metric appropriate or inappropriate. This is the crux of this article. \textit{All methods for mitigating bias and metrics for measuring it must make causal assumptions about the structure of the observed dataset, including ethical assumptions about which causal pathways must be preserved or mitigated.} This is closely related to the recently established fundamental problem of causal fairness analysis (FPCFA) \cite{plecko2022causal}. With this in mind, we now focus on how this problem is relevant in medical imaging.

\section{Medical Mechanisms of (un)Fairness} \label{sec:mechanisms}

In application areas where datasets have a common causal structure and unambiguous sources of bias, the FPCFA may not be insurmountable; it may be possible to develop a standardised toolbox of methods and metrics for mitigating bias (for example, conditional independence testing may be sufficient to identify shifts from a limited set of causal structures  \cite{schrouffDiagnosingFailuresFairness2023}). In medical imaging, however, we deal with a wide range of dataset characteristics stemming from the use of various imaging modalities, different patient populations, clinical tasks, diagnostic processes and workflows, each contributing to the underlying causal processes with different potential sources of bias \cite{bernhardtPotentialSourcesDataset2022}. We illustrate these complications with six clinically inspired examples of dataset biases (one pair for each disparity in Fig. \ref{fig:basicdags}). In each example pair, the causal structures are indistinguishable from observational data alone -- without causal analysis, it would be impossible to mitigate them appropriately.

\subsection*{Prevalence Disparities}

When there is a causal pathway between the sensitive attribute and the prevalence of the disease, we observe \emph{individuals belonging to different subgroups being represented with different prevalence among the targets}.

\begin{figure}[ht]
    \centering
    \includegraphics[width=\linewidth]{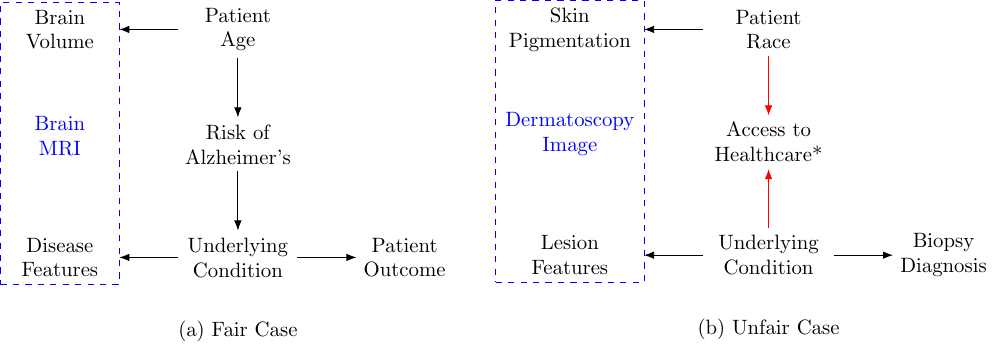}
    \caption{Prevalence disparities in medical imaging. Red arrows represent unfair causal pathways. In our fair example (a), age is a clinically recognised risk factor for Alzheimer's disease. In our unfair example (b), patient race is spuriously correlated with disease prevalence due to disparities in access to healthcare. We mark access to healthcare with * because it is a selection node -- patients may only appear in the dataset if they have access to a dermatology clinic, so the dataset is conditioned upon this variable.}
    \label{fig:representationshift}
\end{figure}

When a dataset exhibits an unfair prevalence disparity, ERM models trained on it will be miscalibrated when deployed in settings where this bias is not present -- leading to under- or over-diagnosis of the condition in one or more groups. We demonstrate an example in Fig. \ref{fig:representationshift}b, involving predicting biopsy outcomes from skin lesion images. In this example, patients from different racial groups may receive disparate access to healthcare \cite{szczepuraAccessHealthCare2005,richardsonAccessHealthHealth2010}, skewing the observed prevalence of the disease at training time. This is a well-studied situation. Approaches for mitigating class imbalance, such as subgroup-aware resampling \cite{liREPAIRRemovingRepresentation2019} and data augmentation \cite{zietlowLevelingComputerVision2022}, may be theoretically sound for tackling prevalence disparities. Similarly, methods for fair representation learning \cite{zemelLearningFairRepresentations2013}, which prevent the model from using sensitive information, may also be appropriate. In particular, adversarial training methods have empirically shown promise in mitigating prevalence disparities on benchmark datasets \cite{alviTurningBlindEye2018,kimLearningNotLearn2019,madrasLearningAdversariallyFair2018,edwardsCensoringRepresentationsAdversary2016}.

In some cases, prevalence disparities may not be unfair. For example, although age is often considered a sensitive attribute, in Fig. \ref{fig:representationshift}a, we demonstrate the task of predicting patient outcomes from brain MRI images. Here, age is a clinically meaningful risk factor for Alzheimer's disease \cite{niccoliAgeingRiskFactor2012,riedelAgeAPOESex2016} -- the correlation between age and disease is not spurious and is not expected to disappear when translating a trained model into clinical practice. We should thus encourage our model to extract age-related features from the image to inform and calibrate its predictions correctly. Methods such as adversarial training on the age attribute would inappropriately worsen performance by forcing models to ignore clinically meaningful information. Suppose our model cannot accurately determine age from the image alone. In that case, it may even be necessary to supply this metadata explicitly at inference time, such as in domain discriminative training \cite{wangFairnessVisualRecognition2020} or decoupled classification \cite{dworkDecoupledClassifiersGroupFair2018}. We draw an analogy, in this case, to human experts, who often use risk factors and other background information beyond the disease features alone to inform their decisions \cite{boykoUseRiskFactors1990}. 

\subsection*{Presentation Disparities}

When there is a causal pathway between the sensitive attribute and the disease-related physiology, we observe \emph{individuals belonging to different subgroups presenting different disease features for the same underlying condition}.

\begin{figure}[ht]
    \centering
    \includegraphics[width=\linewidth]{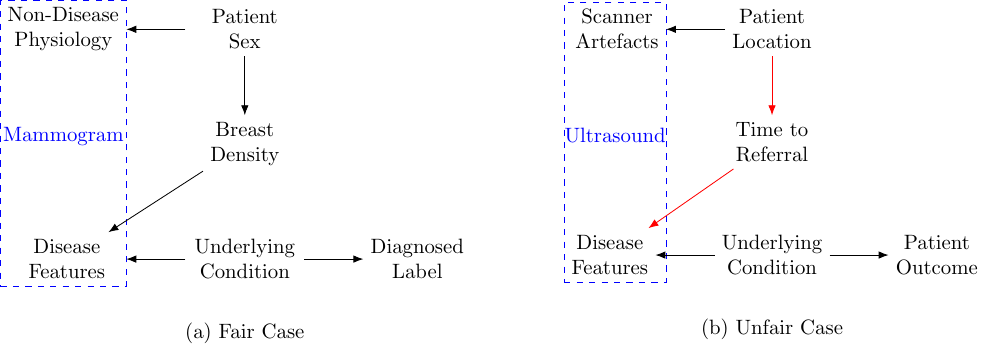}
    \caption{Presentation disparities in medical imaging. Red arrows represent unfair causal pathways. In our fair example (a), breast cancer manifests differently in men and women due to natural differences in breast density. In our unfair example (b), patients in different locations are scanned at different stages due to inconsistent ultrasound referral policies.}
    \label{fig:presentationshift}
\end{figure}

Presentation disparities are complex and often overlooked bias mechanisms and are especially relevant in medical imaging. It is not uncommon for different groups to be scanned with different equipment or to have natural variations in physical characteristics, causing the underlying condition to present differently. In Fig. \ref{fig:presentationshift}b, we see an unfair example in diagnostic ultrasound. Here, patients in different geographic locations are referred for scans at different points in their disease progression due to differences in referral policy \cite{iglehartHealthInsurersMedicalImaging2009,iglehartNewEraMedical2006}. This causes the disease to appear systematically different for groups in each location. This disparity fundamentally alters the mapping from images to targets across groups -- an ERM model trained on this dataset will pick up on location-specific scanner artefacts and will not be transportable to a setting without the bias. Today, it is unclear if it is possible to mitigate unfair presentation disparities in the general case, aside from simply collecting better data. In fact, we postulate that hidden presentation disparities may be a factor behind surprising recent results demonstrating the failure of bias mitigation methods \cite{zietlowLevelingComputerVision2022,wangFairnessVisualRecognition2020}. For instance, Wang et al. \cite{wangFairnessVisualRecognition2020} observe that adversarial training methods worsen task performance on the CIFAR-S benchmark because they are unable to disentangle the sensitive information from the task-relevant information fully. One explanation for this result may be that generating the CIFAR-S dataset (by setting some images to greyscale) introduces a presentation disparity by changing the appearance of the class-specific features across subgroups. 

Fair presentation disparities may occur due to biological differences between groups, such as our example in Fig. \ref{fig:presentationshift}a, showing how tissue density differences may affect breast cancer manifestation in men and women. Like fair prevalence disparities, empirical risk minimisation is appropriate in fair cases of presentation disparities and, again, there may even be situations where we should encourage models to use sensitive information \cite{dworkDecoupledClassifiersGroupFair2018,wangFairnessVisualRecognition2020}. Importantly, when datasets contain presentation disparities, there is no guarantee that the disease should be equally predictable in all groups, so fairness metrics that measure performance disparities across groups are unlikely to be meaningful measures of algorithmic bias when used in isolation.

\subsection*{Annotation Disparities}

When there is a causal pathway between the sensitive attribute and the annotation policy, we observe \emph{individuals belonging to different subgroups being diagnosed, labelled, or annotated with different criteria}.

\begin{figure}[ht]
    \centering
    \includegraphics[width=\linewidth]{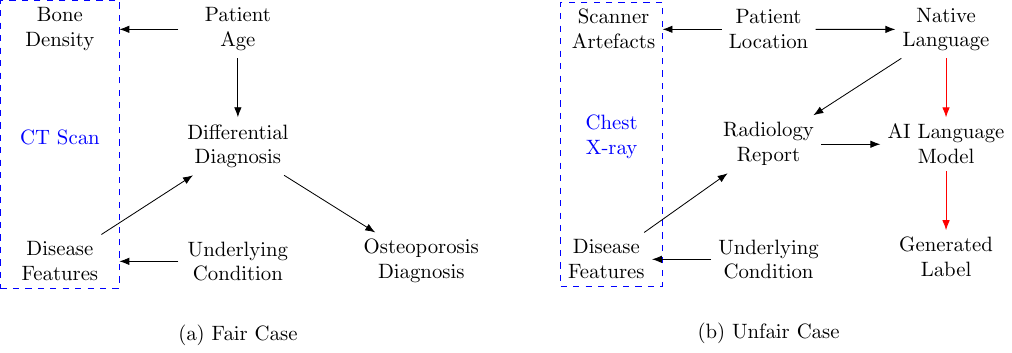}
    \caption{Annotation disparities in medical imaging. Red arrows represent unfair causal pathways. In our fair example (a), patient age is considered a factor in epidemiology-based differential diagnosis of osteoporosis. In our unfair example (b), disease labels are generated automatically from radiology reports with a language model; reports in different languages receive different annotation quality.}
    \label{fig:diagnosisshift}
\end{figure}

We present two examples of annotation disparities in Fig. \ref{fig:diagnosisshift}. These are the most complex causal diagrams in this article, demonstrating how our causal reasoning approach may be applied in messy real-world situations beyond the simple cases of anticausal classification demonstrated so far. Our unfair example in Fig. \ref{fig:diagnosisshift}b is particularly relevant today, showing how the practice of releasing datasets with automatically-generated labels\cite{irvinCheXpertLargeChest2019,johnsonMIMICCXRDeidentifiedPublicly2019} may introduce dataset bias if the label quality varies across subgroups. Consider a hypothetical setting where a natural language model is used to label chest X-ray data from radiology reports. If the language model has inconsistent performance across languages (e.g. because it was trained on a larger English corpus than Spanish), we may see patients in different locations being systematically misdiagnosed based on the language of the radiology reports. Image classifiers trained on such a dataset will pick up on the spurious correlation between location-specific scanner artefacts and label quality and show disparate performance for patients in different locations\cite{jonesRoleSubgroupSeparability2023}. While it may be possible to mitigate unfair annotation disparities by making assumptions on the function class that corrupts the labels \cite{jiangIdentifyingCorrectingLabel2020}, such assumptions may be too strong to justify in practice. Under suspected annotation disparities, it also becomes difficult to interpret evaluation results as there is no reliable ground truth \cite{bernhardtPotentialSourcesDataset2022}. In this case, it may be necessary to relabel the data with a consistent diagnosis policy.

Our fair example of an annotation disparity (Fig. \ref{fig:diagnosisshift}a) involves a setting with differential diagnosis of osteoporosis from CT scans. Since the underlying prevalence of osteoporosis and similar diseases is age-dependent, patients with the same disease features may receive different diagnoses depending on their age due to the clinical practice of differential diagnosis. ERM-trained models will likely reflect this process by picking up on age-related image features such as bone density, which may be desirable, provided we expect to deploy the trained model on an i.i.d. dataset. When deploying such a model, it is crucial to consider whether we expect the deployment hospitals to use the same diagnosis criteria as the training dataset.

\section{Discussion} \label{sec:discussion}

While the study of algorithmic bias is important and has gained significant interest in recent years, underlying dataset biases remain poorly understood. We demonstrated how the causal nature of dataset bias has profound consequences for deep learning algorithms and explored several plausible bias mechanisms in medical datasets. Today, theoretically sound methods exist for tackling a small subset of disparities, but there remains a wide world of underexplored bias mechanisms in medical data. Worse still, even if a method is theoretically appropriate for tackling dataset bias, there is no guarantee that it is easily trainable, and it may yet introduce algorithmic bias if there are issues with training or model selection \cite{zongMEDFAIRBenchmarkingFairness2023,zietlowLevelingComputerVision2022}. Importantly, no bias mitigation method or metric can be successful against all mechanisms of dataset bias; any attempt at mitigating dataset bias must make causal and ethical assumptions about the problem at hand. 

Causal diagrams provide a clear, mathematically principled way of making assumptions about dataset bias explicit. To encourage researchers to consider causality in future problems, we provide a simple three-step framework to aid with causal thinking, displayed in Table \ref{tab:threesteps}. It may be challenging to apply this approach to many problems; however, even attempting to infer causal relationships can provide valuable insights and may highlight important knowledge gaps. As a result, researchers may identify opportunities for engaging with experts from other disciplines. Oftentimes, we do not have perfect knowledge about the data-generating process, and a medical ethicist will likely be needed to evaluate the fairness of causal paths. However, we emphasise that these assumptions must be made (explicitly or implicitly) by \emph{all} methods, so stating them up-front will only improve transparency in machine learning solutions.

\begin{table*}[ht]
\caption{A three-step framework for reasoning about fairness in image analysis. (i) Identify the causal nature of the bias. (ii) Consider the ethics of whether each causal pathway should be preserved or mitigated. (iii) Determine appropriate bias mitigation strategies and metrics which encourage the use of information from fair causal pathways whilst disallowing associations from unfair pathways.}

\centering
\begin{tabular}{@{}lp{0.25\linewidth}p{0.25\linewidth}p{0.25\linewidth}@{}} \toprule
& Presentation Disparities & Prevalence Disparities & Diagnosis Disparities \\ \midrule
(i) & \multicolumn{3}{l}{\textbf{Identification of the bias:}} \\
\vspace{2\baselineskip}\\

& Does the disease present differently in different subgroups? & Is the disease prevalence uneven across subgroups? & Are similar individuals in different subgroups assigned different labels? \\ \midrule


(ii) & \multicolumn{3}{l}{\textbf{Ethical consideration of the pathways:}} \\
\vspace{2\baselineskip}\\
& \multicolumn{3}{c}{Is the disparity expected to be preserved in deployment?} \\
& \multicolumn{3}{c}{Is there a known biological mechanism causing the difference?} \\
& \multicolumn{3}{c}{Would a medical ethicist consider it appropriate to use sensitive information\cite{mccraddenWhatFairFair2023}?}\\ \midrule

(iii) & \multicolumn{3}{l}{\textbf{Metrics and mitigation in the fair case:}} \\
\vspace{2\baselineskip}\\

& \multicolumn{3}{c}{Empirical Risk Minimisation is likely appropriate.} \\
& \multicolumn{3}{c}{Measure algorithmic bias with bias-preserving metrics \cite{wachterBiasPreservationMachine2021}.}\\
& \multicolumn{3}{c}{ Consider leveraging sensitive information at inference time \cite{dworkDecoupledClassifiersGroupFair2018}.} \\

\vspace{2\baselineskip}\\

& \multicolumn{3}{l}{\textbf{Metrics and mitigation in the unfair case:}} \\
\vspace{2\baselineskip}\\

& Unclear if any of today's methods are appropriate. Consider collecting or generating debiased data \cite{noriega-camperoActiveFairnessAlgorithmic2019,vanbreugelDECAFGeneratingFair2021}. & Collect more data; mitigate class imbalance \cite{liREPAIRRemovingRepresentation2019}; learn debiased representations \cite{alviTurningBlindEye2018,kimLearningNotLearn2019, madrasLearningAdversariallyFair2018,edwardsCensoringRepresentationsAdversary2016}. &  Only fixable with strong assumptions about the bias \cite{jiangIdentifyingCorrectingLabel2020}. Consider relabelling data consistently.\\

\vspace{2\baselineskip}\\

& Maximise subgroup-wise performance. Metrics are likely not comparable across groups. & Maximise subgroup-wise performance and monitor bias-transforming metrics \cite{wachterBiasPreservationMachine2021}. & Unclear whether annotation disparities are detectable without a ground truth \cite{bernhardtPotentialSourcesDataset2022}. \\

\bottomrule
\end{tabular}
\label{tab:threesteps}
\end{table*}

The bias mechanisms we explored are particularly relevant to medical imaging but are not an exhaustive list. Our proposed mechanisms focus on the common case of anticausal classification and contain one causal pathway connecting the sensitive attribute to the primary task; however, real-world datasets may simultaneously contain multiple fair and unfair paths or have a different underlying causal structure (e.g. image segmentation may be better modelled as ${Z \rightarrow X \rightarrow Y}$). Today, there is little work combating compound bias mechanisms, and we are unaware of any image analysis methods that can simultaneously handle fair and unfair bias mechanisms. Meanwhile, we call for medical imaging datasets to be more transparent regarding the processes involved in collection and curation \cite{gebruDatasheetsDatasets2021,pushkarnaDataCardsPurposeful2022,mitchellModelCardsModel2019,liuMedicalAlgorithmicAudit2022,mccraddenWhatFairFair2023} -- without such knowledge, we cannot make informed assumptions about the nature of dataset biases.  One of the simplest yet most effective methods for bias mitigation remains the active collection of unbiased data \cite{noriega-camperoActiveFairnessAlgorithmic2019}, and a promising future direction may be to apply counterfactual generative models \cite{pawlowskiDeepStructuralCausal2020,ribeiroHighFidelityImage2023} for leveraging causal assumptions to generate debiased synthetic data \cite{vanbreugelDECAFGeneratingFair2021}. 

There remains a great opportunity for machine learning to improve patient outcomes in healthcare. Still, the tendency of today's methods to produce unfair and inequitable predictions will continue to restrict true progress in the field. Causal reasoning provides a principled formulation of the problems we face and should play a role in future analyses of the issues.

\subsection*{Acknowledgements}

C.J. is supported by Microsoft Research and EPSRC through the Microsoft PhD Scholarship Programme. B.G. received support from the Royal Academy of Engineering as part of his Kheiron/RAEng Research Chair.

\bibliographystyle{naturemag}
\bibliography{references}

\newpage

\appendix
\section{Primer on Graphical  Causality} \label{sec:causalprimer}

Causal reasoning
\cite{pearlCausalityModelsReasoning2011,petersElementsCausalInference2017}
allows us to examine the relationships that give rise to real-world datasets. At
the basic level, a set of random variables $\{A, B, C, D\}$ (say, age, bladder
cancer, cigarette consumption, and deafness, respectively) may be \emph{associated}
through statistical correlations. However, we can only say that $C$ causes $B$
if \emph{intervening} on the value of $C$ changes the distribution of $B$. For
example, smoking more cigarettes may increase risk of bladder cancer, but the
reverse is not true -- people who develop cancer do not automatically become
smokers! Similarly, ageing may be a cause of bladder cancer and deafness but
be causally unrelated to cigarette consumption. This distinction between
association and intervention is formalised in Pearl's causal hierarchy
\cite{pearlCausalityModelsReasoning2011}, defining three successive levels
of causal reasoning, each requiring knowledge that provably cannot be
gleaned from previous levels \cite{bareinboim2022pearl}. While observing
associations lies on the first level, predicting effects of interventions requires
a causal model of the world, so is on the second level. Without such a model,
observational data alone provides no information about the effects of interventions; hence the statistician's refrain, `correlation does not imply causation'.
A final level of causal reasoning, \emph{counterfactual} inference, also exists,
where we imagine alternate worlds where variables were to be
different. Counterfactual calculations apply on the individual level, unlike
interventions which consider average effects over populations. With
counterfactual inference, we are able to answer questions such as ``if I had
smoked a pack of cigarettes a day, would I have developed bladder cancer at age
25?". This is the most powerful level of causal reasoning and requires the strongest
assumptions. By combining a causal model with knowledge about what has
already happened in one world, counterfactual inference considers what might
have happened in another.

\begin{figure}[ht]
    \centering
    \begin{subfigure}[b]{0.4\linewidth}
        \centering
        \includegraphics{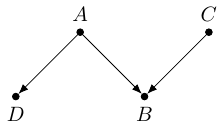}
    \end{subfigure}
    \begin{subfigure}[b]{0.4\linewidth}
        \centering
        \begin{align*}
            A & \perp C             \\
            B & \not\perp D         \\
            A & \not\perp C \mid B \\
            B & \perp D \mid A
        \end{align*}
    \end{subfigure}
    \caption{ Left: a causal diagram encoding possible relationships between age
        ($A$), bladder cancer ($B$), cigarette consumption ($C$), and
        deafness ($D$). Right: independence relationships implied by the DAG.}
    \label{fig:causalprimer}
\end{figure}

In the language of causality, we draw causal diagrams as directed
acyclic graphs (DAGs) and can visually reason about associations, interventions,
and counterfactuals. Fig. \ref{fig:causalprimer} demonstrates a causal diagram
for our hypothetical smoking example. In causal diagrams, nodes are random variables and arrows
represent cause-effect relationships, where $A \rightarrow B$ implies that $A$ causes $B$. Note that the absence of arrows also encodes useful information because the value of each node is, by definition, some function of its parents and uncorrelated exogenous noise only. We can construct causal diagrams to represent the assumed data-generating process of real-world datasets, in which case the DAG implies a set of conditional independence relationships in the observational distribution (in general, there may be many DAGs which imply the same set of relationships, this is referred to as a Markov equivalence class). We can infer these relationships through
the d-separation criterion \cite{vermaCausalNetworksSemantics1990,pearlIdentifyingIndependenciesCausal1996}, which is best explained through four simple rules:

\begin{itemize}
    \item $\mathcal{X} \perp \mathcal{Y} \mid \mathcal{Z}$ if the sets of variables $\mathcal{X}$ and $\mathcal{Y}$ are d-separated by the set $\mathcal{Z}$.
    
    \item Two variables $X$ and $Y$ are d-connected (i.e. not d-separated) if there is an unblocked path between them, where a path is a direction-agnostic sequence of edges. 
    
    \item Connected paths are blocked by colliders (nodes with at least two parents).

    \item Conditioning on a variable $Z$ blocks connected paths through it. If $Z$ is a collider or descendant of a collider, conditioning unblocks it instead.

\end{itemize}

Using our example in Fig. \ref{fig:causalprimer}, we can see there are
no unblocked pathways linking age to cigarette consumption; hence we can declare
that the variables are independent in the observational distribution (i.e. $A \perp C$). Importantly, since $B$ is a collider, conditioning on $B$ opens the pathway from
$A$ to $C$. In practice, this means that a statistician studying only people
with bladder cancer would incorrectly conclude that there is an association between
age and cigarette consumption (i.e. $A \not\perp C \mid B$). This is an example of a
selection bias \cite{hernanStructuralApproachSelection2004} and demonstrates how
causal reasoning can help us avoid common pitfalls in statistical inference.
Similarly, $A$ is a \emph{confounder}, a cause of two other variables ($D$ and $B$
in this case). In general, confounders should be controlled for according to the
backdoor criterion \cite{pearlCausalDiagramsEmpirical1995,rosenbaumCentralRolePropensity1983}. 
A statistician studying this data without controlling for age may incorrectly infer that the 
spurious association between deafness and bladder cancer is clinically meaningful (i.e. $B \perp D \mid A$).

Causal inference is also useful for studying the transportability (or `generalisation') of statistical models. In supervised machine learning tasks, we train models which learn a mapping ${P(Y \mid X)}$ from an observed training dataset. While statistical learning theory gives us guarantees about model performance in i.i.d. settings \cite{vapnikOverviewStatisticalLearning1999}, we often care about whether performance will degrade when deployed in the real world, where the data-generating process is likely to differ from the training dataset. Pearl and Bareinboim \cite{pearlTransportabilityCausalStatistical2011} formalise this concept of causal transportability, and in visual recognition, Jiang and Veitch \cite{jiangIdentifyingCorrectingLabel2020}, and Mao et al. \cite{maoCausalTransportabilityVisual2022} apply it to anticausal and causal tasks, respectively. While the rules of transportability are too involved to recap in this article, intuitively, our models should learn true causal mechanisms, which will remain consistent across domains. In contrast, we wish to resist exploiting spurious correlations, which will likely shift from training time to deployment\cite{arjovskyInvariantRiskMinimization2019}.

\end{document}